\documentclass[oneside,onecolumn,letter]{article}
\usepackage[citestyle=numeric-comp,maxbibnames=8, sorting=none, giveninits=true]{biblatex}

\usepackage{amsmath,amsfonts,bm}

\def\eqref#1{equation~\ref{#1}}

\def\1{\bm{1}}

\DeclareMathAlphabet{\mathsfit}{\encodingdefault}{\sfdefault}{m}{sl}
\SetMathAlphabet{\mathsfit}{bold}{\encodingdefault}{\sfdefault}{bx}{n}

\usepackage[hidelinks]{hyperref}
\usepackage{url}

\usepackage[ruled]{algorithm2e}
\usepackage[many]{tcolorbox}
\usepackage{graphicx}
\usepackage{booktabs}
\usepackage{multirow}
\usepackage{enumitem}
\usepackage[normalem]{ulem}
\usepackage{subcaption}
\usepackage{microtype}
\usepackage{url}
\usepackage{placeins}   
\usepackage[acronym,automake]{glossaries}
\makeglossaries
\renewcommand{\glossarysection}[2][]{}
\usepackage{authblk}
\usepackage[margin=1in]{geometry}
\usepackage[parfill]{parskip}
\usepackage{array}
\newcolumntype{L}[1]{>{\raggedright\let\newline\\\arraybackslash\hspace{0pt}}m{#1}}
\newcolumntype{C}[1]{>{\centering\let\newline\\\arraybackslash\hspace{0pt}}m{#1}}
\newcolumntype{R}[1]{>{\raggedleft\let\newline\\\arraybackslash\hspace{0pt}}m{#1}}

\makeatletter
\renewcommand\AB@affilsepx{, \protect\Affilfont}
\makeatother


\newcommand\blfootnote[1]{%
  \begingroup
  \renewcommand\thefootnote{}\footnote{#1}%
  \addtocounter{footnote}{-1}%
  \endgroup
}

\title{TabDistill: Distilling Transformers into Neural Nets for \\ Few-Shot Tabular Classification}

\author{Pasan Dissanayake$^*$}
\author{Sanghamitra Dutta}
\affil{University of Maryland, College Park}
\date{}

\newacronym[
    longplural={Gradient Boosted Decision Trees},
    shortplural={GBDTs}
]{GBDT}{GBDT}{Gradient Boosted Decision Tree}

\newacronym[
    longplural={Large Language Models},
    shortplural={LLMs}
]{LLM}{LLM}{Large Language Model}

\newacronym[
    longplural={Multi-Layer Perceptron},
    shortplural={MLPs}
]{MLP}{MLP}{Multi-Layer Perceptron}

\newacronym{ROC-AUC}{ROC-AUC}{Receiver Operating Characteristic Area Under the Curve}

\def\methodName{TabDistill} 

\begin{document}

\maketitle

\begin{abstract}
Transformer-based models have shown promising performance on tabular data compared to their classical counterparts such as neural networks and \glspl{GBDT} in scenarios with limited training data. They utilize their pre-trained knowledge to adapt to new domains, achieving commendable performance with only a few training examples, also called the few-shot regime. However, the performance gain in the few-shot regime comes at the expense of significantly increased complexity and number of parameters. To circumvent this trade-off, we introduce \methodName{}, a new strategy to distill the pre-trained knowledge in complex transformer-based models into simpler neural networks for effectively classifying tabular data. Our framework yields the best of both worlds: being parameter-efficient while performing well with limited training data. The distilled neural networks surpass classical baselines such as regular neural networks, XGBoost and logistic regression under equal training data, and in some cases, even the original transformer-based models that they were distilled from.
\end{abstract}

\blfootnote{$^*$Corresponding author. Email: \{pasand, sanghamd\}@umd.edu}

\glsresetall

\section{Introduction}
\label{sec_introduction}

Tabular data plays a central role in high-stakes applications, ranging from finance and healthcare, to manufacturing and weather prediction \cite{shwartz2022deepLearningNotAllYouNeed,van2024tabular}. However, the scarcity of labeled data can limit the application of machine learning in these domains, e.g., some diseases are extremely rare, or certain natural phenomena occur once in centuries~\cite{hegselmann2023TabLLM, nam2023stunt,hamman2024predictionConsistency}. In financial applications, annotating data can be expensive, and suffer from issues such as subjectivity, mislabeling, lack of consensus, and also data imbalances where only the data of accepted applicants may be available but not the rejected group~\cite{crook2004does}. Thus, tabular classification models that perform well under limited training data, also called the few-shot regime, are of immense interest.

Recently, transformer-based models have been shown to surpass classical approaches such as neural networks and \glspl{GBDT} in the few-shot regime when the number of training examples is significantly small~\cite{hollmann2023TabPFN, hegselmann2023TabLLM, jayawardhana2025PFNBoost, dinh2022lift}. While \glspl{GBDT} such as XGBoost \cite{chen2016xgboost}, CatBoost \cite{prokhorenkova2018catboost}, and LightGBM \cite{ke2017lightgbm} have long been the state-of-the-art for tabular classification when there is sufficient labeled data for training~\cite{shwartz2022deepLearningNotAllYouNeed, grinsztajn2022WhyDoTreeBasedModels}, transformer-based models instead exploit their pre-trained knowledge to achieve improved performance in the few-shot regime. However, the performance gain in few-shot regime comes at the expense of efficiency. Transformer-based models are extremely complex (millions or billions of parameters) in comparison to traditional neural networks and \glspl{GBDT}, requiring massive compute, energy, and time during inference. To be able to cater to applications across varying levels of infrastructure, it is usually desirable that the deployed models are parameter-efficient and scalable. In this work, our key question is: \emph{Can we achieve the best of both worlds, i.e., being parameter-efficient while also performing well with limited training data?}

\begin{figure}[ht]
    \centering
    \begin{subfigure}[h]{\textwidth}
        \centering
        \includegraphics[width=\textwidth]{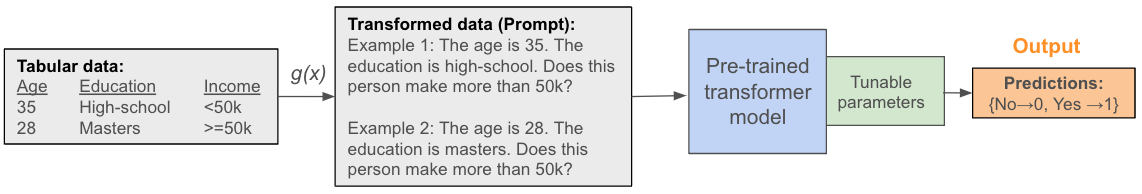} 
        \caption{The TabLLM framework. Tabular data is first converted to a natural language string using a serialization technique (denoted by $g(x)$). The serialized text is given as the input to the \gls{LLM} and a prediction is directly generated as the output. Fine-tuning the LLM can improve classification performance.}
        \label{fig_tabllm}
    \end{subfigure}
    \begin{subfigure}[h]{\textwidth}
        \centering
        \includegraphics[width=\textwidth]{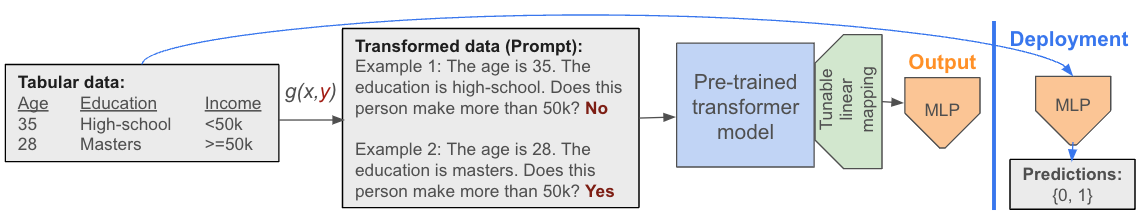}
        \caption{Our \methodName{} framework. Similar to TabLLM, the serialized text is given as the input to the \gls{LLM}. However, in contrast to TabLLM, an \gls{MLP} is generated as the \textbf{final} output of the transformer model. Only this \gls{MLP} is deployed for making predictions on real-world data. Fine-tuning the LLM gives an improved \gls{MLP}.}
        \label{fig_tabdistillOverview}
    \end{subfigure}
    \caption{Comparison of TabLLM and \methodName{} frameworks. The tunable parameters which are fine-tuned during training in each framework are depicted in green. The example dataset contains \texttt{Age} and \texttt{Education} as features. The target is to predict whether the \texttt{Income} is $>=50$k or not.}
    \label{fig_transformerClassifiers}
\end{figure}

Toward answering this question, we propose \methodName{}, a novel framework to distill the classification capabilities of pre-trained transformers into neural networks for few-shot tabular classification (also see Figure~\ref{fig_transformerClassifiers}). We draw inspiration from the image domain where transformer-based models have been found to be good hypernetworks for generating neural networks to implicitly represent images~\cite{Chen2022TransINR, gu2025enhancingFoundationModelsHypernetworks}. \textit{Succinctly, \methodName{} incorporates the pre-trained knowledge of a transformer-based model (the base model) into a neural network by fine-tuning the transformer to infer its weights}. We assume that the base model contains an informative intermediate representation (for example, the encoder output of TabPFN \cite{hollmann2023TabPFN} or BERT \cite{devlin2019bert}, encoder-decoder type language models such as BART \cite{lewis2019bart} and BigScience T0 series \cite{sanh2021t0Models} etc.). \methodName{} learns a linear map for projecting the intermediate representation provided by the base model into the parameter space of the neural network, by fine-tuning using the cross-entropy loss of the resultant classifier, and achieves both data efficiency and parameter efficiency. We employ a novel permutation-based training technique to avoid overfitting the model to the extremely small number of training examples.

Our experiments span over five tabular datasets and two base models. We compare the \methodName{} framework with 5 baselines, including 3 classical models and the 2 base models. Experimental results indicate that the neural network distilled using the proposed framework exceeds the classical baselines in performance, particularly in the very-few-shot regime (when the number of training examples is less than 10). Interestingly, under some settings, the distilled neural network exceeds the performance of the base transformer-based model that it was distilled from. 

In summary, our main contributions can be listed as follows:
\begin{itemize}[itemsep=0pt, leftmargin=*, topsep=0pt]
    \item {\bf Propose \methodName{}, a novel framework to distill transformers into neural networks.} 
    We introduce a new strategy that we call \methodName{} to extract the performance of transformers into a much more efficient \gls{MLP}. Accordingly, the framework has additional advantages of the resulting model being differentiable and more easily explainable.
    \item  {\bf Instantiate the framework with two transformer-based models.} We instantiate the distillation framework with two transformer-based models Bigscience T0pp \cite{sanh2021t0Models} and the more recent TabPFN \cite{hollmann2023TabPFN}, which have ${\sim}11$B and ${\sim}11$M parameters respectively. We distill these base models into significantly simpler neural networks with ${\sim}1000$ parameters.
    \item {\bf Experimental validation.} We conduct experiments on five tabular datasets (Bank \cite{bankDataset}, Blood \cite{bloodDataset}, Calhousing \cite{calhousingDataset}, Heart \cite{heartDataset}, and Income \cite{incomeDataset}) and five baselines (\gls{MLP}, logistic regression, XGBoost, and the two base models TabPFN and T0pp). The distilled \gls{MLP} obtained using our \methodName{} surpasses the classical baselines in the few-shot regime under equal training data, and in some cases, even the original transformer-based models that they were extracted from.
\end{itemize}

\subsection{Related Works}

\textbf{Classical algorithms for tabular data.} Despite the success of deep learning in various other domains, classical machine learning algorithms such as logistic regression and \gls{GBDT} methods such as XGBoost \cite{chen2016xgboost}, LightGBM \cite{ke2017lightgbm} and CatBoost \cite{prokhorenkova2018catboost} have been dominating the domain of tabular datasets \cite{shwartz2022deepLearningNotAllYouNeed}. 
In \cite{grinsztajn2022WhyDoTreeBasedModels}, the authors point out that the deep learning models struggle on tabular datasets mostly due to difficulties in learning irregular patterns of the target function. Multiple works have focused on overcoming such difficulties and adapting neural networks for tabular datasets (See \cite{gorishniy2024tabM, arik2021tabnet, popov2019node} and references therein). However, given the fact that these classical models are trained from scratch for a given dataset, their performance degrades significantly in the few-shot regime \cite{hegselmann2023TabLLM}. 

\textbf{Transformer-based models for tabular data.} Transformer-based models have seen promising performance gains within the tabular data domain. A multitude of works employ the transformer as a way to model complex interactions between features of a tabular dataset. SAINT \cite{somepalli2022saint} uses an attention mechanism across rows as well as columns to better learn the structures within data. It also incorporates a self-supervised pre-training method for situations where the labels are scarce. In \cite{hollmann2023TabPFN}, the authors train a transformer from scratch on a massive collection of synthetic tabular datasets sampled from a causal mechanism. The trained transformer TabPFN can then be used to predict new tabular tasks with no additional training. \textcite{dinh2022lift} shows that \glspl{LLM} can be fine-tuned using text serializations of tabular data for both classification and regression tasks. In an attempt to leverage the pre-trained knowledge of a \gls{LLM} for tabular data classification, \textcite{hegselmann2023TabLLM} fine-tune models from BigScience T0 series \cite{sanh2021t0Models} to achieve remarkable performance in the few-shot regime. \textcite{jayawardhana2025PFNBoost} propose PFN-Boost and LLM-Boost techniques where a pre-trained transformer is incorporated as the initial weak classifier of a \gls{GBDT} ensemble. \textcite{yan2024makingLMsGreat} introduce a novel framework for tabular data prediction using \glspl{LLM} which involves representing numerical feature values as relative magnitude tokens. They pre-train an \gls{LLM} on a large number of tabular datasets and use it for downstream tasks with an attached classification head. \textcite{yin2020tabert} propose a framework for learning a joint representation of natural language queries and tabular data which is useful in multiple downstream tasks. The framework involves first discovering a useful subset of rows from the table and serializing, and then combining the serialized rows with the query to generate a shared representation. These works underscore the capabilities of transformer-based models in modeling and understanding tabular data. However, the performance gain of these methods is offset by the increased model complexity and resource consumption particularly during inference. Moreover, the increased complexity causes difficulty in assessing reliability, for instance, model multiplicity \cite{hamman2024predictionConsistency}. Our method focuses on mitigating these limitations by distilling the transformer into an \gls{MLP}.

\textbf{Meta-Learning and hypernetworks.}
Meta-learning is the process of learning to generalize to unseen tasks by observing few examples corresponding to each task~\cite{vilalta2002metaLearningPerspective}. Transformers are known to be good at meta learning, particularly in the form of in-context learning~\cite{kirsch2022metaLearningTransformers}. Hypernetworks are closely related meta-learning, in the sense that they predict parameters for other machine learning models by observing a few samples from the task at hand. Transformers have been used as hypernetworks in computer vision applications, specifically for generating implicit neural representations~\cite{Chen2022TransINR, gu2025enhancingFoundationModelsHypernetworks}. In \cite{Chen2022TransINR}, a transformer is trained from scratch to predict weights of a neural network which represents an image or a 3D scene. Another work~\cite{gu2025enhancingFoundationModelsHypernetworks} exploits the pre-trained knowledge of a transformer-based foundation model for the same task. Both the works append additional placeholder tokens to the input for predicting the neural network weights. In contrast, our framework directly maps the embedding space to neural network parameters. MotherNet~\cite{mueller2025mothernet} is a hyper-network based on a TabPFN-like model which provides a neural network classifier in a single forward pass with in-context learning, given a set of tabular training data. It is pre-trained on a large number of synthetic tabular data. After the initial pre-training phase (i.e., the meta-learning phase), MotherNet does not need any additional training to adapt to an unseen dataset. In contrast to our work, the focus is not on the few-shot regime, and the architecture of the generated neural network is fixed irrespective of the actual dataset and its features. 

\textbf{Knowledge distillation.} Knowledge distillation~\cite{gouKDSurvey,sucholutsky2023getting,hintonDistillation, romeroFitnets,dissanayakequantifying,xu2024survey,liang2023less,hamman2025few} is another popular paradigm that trains a smaller student model from a larger teacher model (often from the same model family) by directly aligning the student's outputs with the teacher's outputs using a loss function. Our setup is distinctly different since we extract a neural network out of a transformer-based model without any loss-function-based alignment.

\section{\methodName: Distilling transformers into neural networks}
\label{sec_methodology}

In this section, we first discuss our proposed \methodName{} framework along with the training procedure and possible methods for hyperparameter tuning. We then elaborate on two example instantiations of the framework using two popular transformer-based models for few-shot tabular data classification, namely, TabPFN \cite{hollmann2023TabPFN} and TabLLM \cite{hegselmann2023TabLLM}. 

\subsection{Notation and problem setup}

Let $\mathcal{D}_N=\left\{(x_n,y_n), x_n\in \mathcal{X}, y_n\in\{0, 1\}, n=1,\dots,N\right\}$ be a small tabular dataset for binary classification with $d$ features (usually pre-processed, e.g., categorical features are one-hot encoded) and $N$ datapoints ($N\sim 10$). Our focus is on transformer-based models capable of classifying instances $x\in\mathcal{X}$ of tabular datasets. To this end, \glspl{LLM} have been adapted as few-shot classifiers through parameter-efficient fine-tuning~\cite{hegselmann2023TabLLM}. To perform classification using an \gls{LLM}, the tabular data instance $x$ must first be transformed into a natural language string (denoted by $s\in\mathcal{S}$ where $\mathcal{S}$ is the space of all possible strings within a given length). \cite{hegselmann2023TabLLM} studies a wide array of techniques for converting rows of a tabular dataset into text, known as ``serialization techniques". These techniques include using a fixed text template such as ``\texttt{The \textless column\textunderscore name> is \textless value>}" and using a list template of the form ``\texttt{\textless column\textunderscore name>: \textless value>}." We denote such a serialization by $g(x):\mathcal{X}\rightarrow \mathcal{S}$. 

The text output of the \gls{LLM} can be converted to a binary class prediction using a similar technique (for example, $\texttt{Yes}\rightarrow 1$, and $\texttt{No} \rightarrow 0$). We abstract out this mapping and denote the transformer model by $f(s):\mathcal{S}\rightarrow \{0,1\}$. Note that with this setup, a meaningful classification can be carried out by predicting $\hat{y} = f\left(g(x))\right)$. For brevity, let $g(x,y)$ represent a similar transform applied to both the features and the label together, and $g(\mathcal{D}_N)$ represent the concatenation of all the strings $g(x_n, y_n)$ corresponding to each $(x_n, y_n)\in \mathcal{D}_N$. See Figure \ref{fig_transformerClassifiers} for an example application of a text template. It is worth noting that TabPFN \cite{hollmann2023TabPFN} takes the tabular feature values themselves as the input and hence, $\mathcal{S}=\mathcal{X}$ and $g(x)$ in this case is the identity function, i.e., $g(x)=x$.

\textit{Our goal is to use the pre-trained knowledge of the complex transformer-based model $f$ to generate a much simpler \gls{MLP} $h_\theta(x):\mathcal{X}\rightarrow \{0, 1\}$ with parameters $\theta\in \Theta$ that can classify $x\in\mathcal{X}$.} The intuition is that the pre-trained knowledge of $f$ will assist in generating $h_\theta$ effectively in a few-shot setting (i.e., when $N$ is very small). We consider the complex model $f(s)$ as consisting of two major components: an encoder $f_E(s):\mathcal{S}\rightarrow \mathcal{Z}$ and a decoder $f_D(z):\mathcal{Z} \rightarrow \{0, 1\}$, where $\mathcal{Z}$ is an embedding space. This is the case for the transformer-based models used in TabLLM \cite{hegselmann2023TabLLM} and TabPFN \cite{hollmann2023TabPFN} as well as popular \glspl{LLM} such as BERT \cite{devlin2019bert}, BART \cite{lewis2019bart}, and T5 \cite{raffel2020t5}.

The \gls{MLP} $h_\theta(x)$ has the following architecture. Let $\texttt{ReLU}(u)$ denote the ReLU activation function \cite{glorot2011relu}. With the hyperparameters $R$ and $L$ denoting the number of layers and the width of the hidden layers, respectively, $h_\theta(x)$ is defined as:
\begin{equation}
    h_\theta(x) = \texttt{ReLU}\left(   W_R \texttt{ReLU}\left(\cdots  \texttt{ReLU}\left(W_2 \texttt{ReLU}\left(  W_1 x + b_1  \right) + b_2 \right) \cdots\right) + b_R  \right)
\end{equation}
where $W_i$ and $b_i$ $(i=1,\dots,R)$ are the weights and the biases of each linear layer. The parameter $\theta$ denotes the combination of all such weights and biases, i.e., $\theta = (W_1, b_1, W_2, b_2, \dots, W_R, b_R)$ and hence, $\text{dim}(\Theta)$ is equal to the total number of tunable parameters in $h_\theta$ determined by $d, R$ and $L$. The first matrix $W_1 \in \mathbb{R}^{L \times d}$, where $d$ is the input dimension. All the intermediate layers have $W_i\in\mathbb{R}^{L\times L}$ for $i=2,\dots,R-1$ and the final layer has $W_R \in \mathbb{R}^{L \times 2}$ for binary classification. For all $i=1,\dots,R$, $b_i\in \mathbb{R}$. The output logits $h_\theta(x)$ can be normalized by applying a Softmax function $\sigma(\cdot)$ to get the final class probability predictions. If desired, one can also choose different dimensions for each weight matrix rather than a fixed $L$.

\subsection{Our proposed \methodName{} framework}

\begin{figure}[ht]
    \centering
    \includegraphics[height=4.8cm]{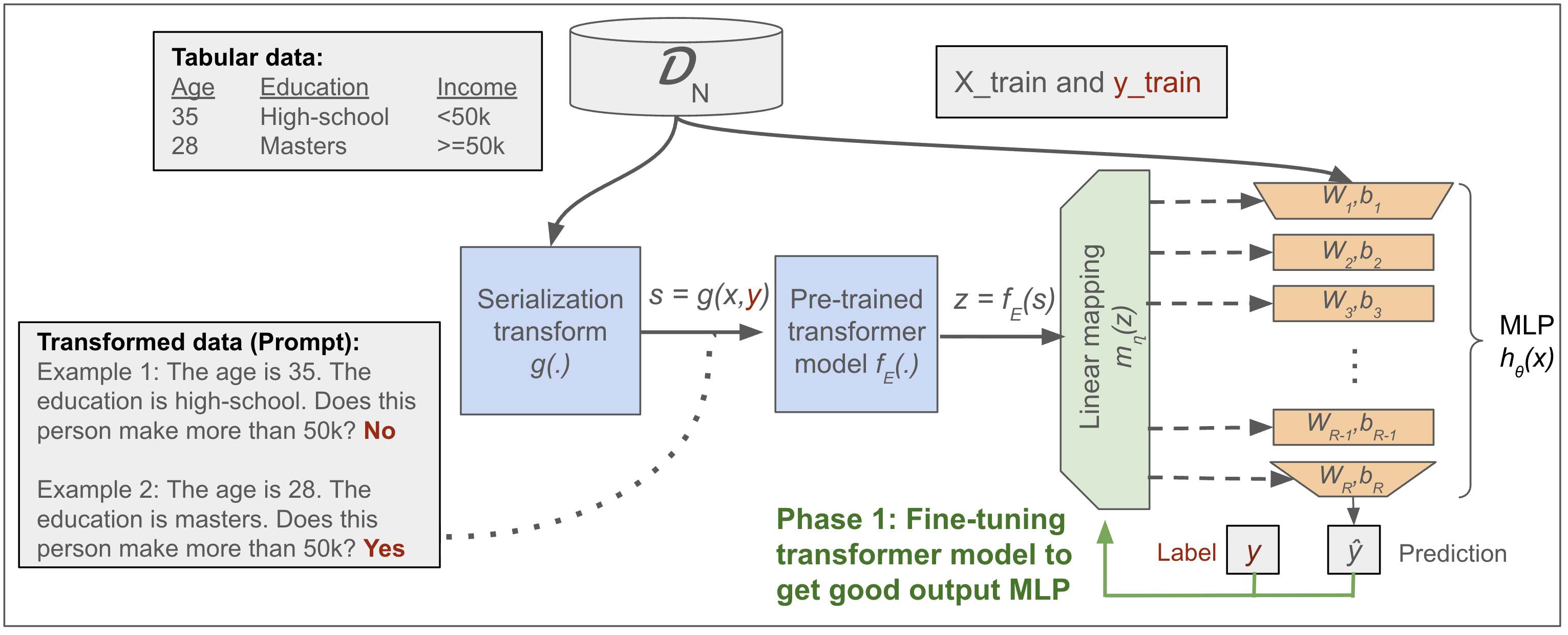}
      \includegraphics[height=4.8cm]{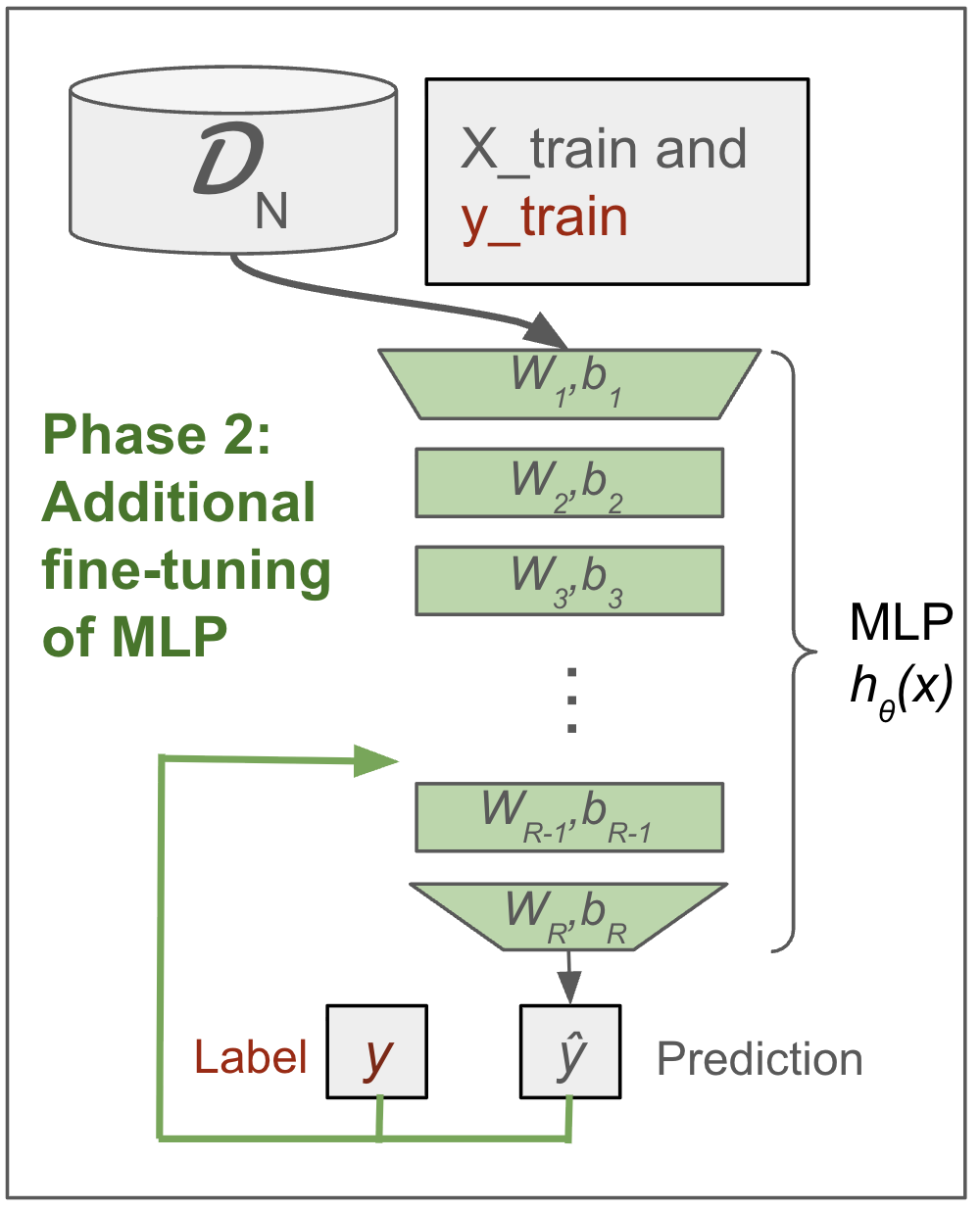}
    \caption{\methodName{} framework. In Phase 1 (left), the tunable parameters of the transformer model (the linear mapping $m_\eta(z)$) is fine-tuned, as depicted in green. The resultant output \gls{MLP} $h_\theta$ is depicted in amber. When T0pp is used as the base model $f$, a text serialization $g(x, y)$ is applied as shown in the figure. When TabPFN is used as the base model, $g(x, y)$ becomes the identity function. In Phase 2 (right), the MLP may be further fine-tuned if desired, as depicted in green.}
    \label{fig_framework}
\end{figure}
\textbf{Phase 1: Fine-tuning the base transformer model.} The distillation is achieved by using the encoder $f_E$ of the complex model for inferring the weights of the \gls{MLP} $h_\theta$. We learn a linear mapping $m_\eta(z):\mathcal{Z}\rightarrow \Theta$ parameterized by $\eta$ such that $\theta=m_\eta(f_E(g(\mathcal{D}_N)))$ results in a useful classifier $h_{\theta}$. We use a simple normalized linear layer as the mapping function, defined as $m_\eta(z) = \texttt{LayerNorm}(Az + b)$, where $A \in \mathbb{R}^{\text{dim}(\Theta) \times \text{dim}(\mathcal{Z})}$, $b\in \mathbb{R}^{\text{dim}(\Theta)}$ and $\eta=(A, b)$. During Phase 1, the fine-tuning loss $\mathcal{L}(\eta; \mathcal{D}_N)$ is computed as follows:
\begin{enumerate}[leftmargin=*,itemsep=0pt,topsep=0pt]
    \item Create a combined serialized input $g(\mathcal{D}_N)$ for the complex model $f$ using the training data $\mathcal{D}_N$.
    \item Prompt the complex model encoder to get the embeddings $z=f_{E}(g(\mathcal{D}_N))$.
    \item Infer the parameters $\theta=m_\eta(z)$ for the \gls{MLP} $h_\theta$.
    \item Compute the loss $\mathcal{L}(\eta; \mathcal{D}_N)$ as the cross entropy loss of the classifier $h_\theta$. 
\end{enumerate}
Ultimately, the fine-tuning loss function can be written as:
\begin{equation}
\label{eq_trainingLoss}
    \mathcal{L}(\eta; \mathcal{D}_N) = \sum_{n=1}^N y_n \log\Big(\sigma(h_\theta(x_n))[[1]]\Big) + (1-y_n) \log\Big(\sigma(h_\theta(x_n))[[0]]\Big)
\end{equation}
with $\theta={m_\eta(f_E(g(\mathcal{D}_N)))}$ and the indexing $[[c]]$ for $c\in \{0, 1\}$ indicates the corresponding predicted class probabilities. Note that the parameters of the complex model $f$ do not undergo any modifications during the fine-tuning phase since this is a form of parameter-efficient fine-tuning.

\textbf{Phase 2: Additional fine-tuning of the MLP.} The distilled \gls{MLP} $h_\theta$ is extracted by prompting $f_E$ with the same training dataset $\mathcal{D}_N$. One can further fine-tune $h_\theta$ for $K$ additional epochs on $\mathcal{D}_N$. During inference, the predictions are made using $h_\theta$, similar to any ordinary \gls{MLP}. The complex model $f$ is no longer involved in the inference phase after the initial extraction of $h_\theta$.

\textbf{Nature of the input prompt:} The same small training set $\mathcal{D}_N$ (or a subset of $\mathcal{D}_N$) is used for two tasks during Phase 1: First, a serialized/transformed version of $\mathcal{D}_N$ (i.e., $g(\mathcal{D}_N)$) is used to prompt the base model $f$ to retrieve $h_\theta$. Then, $\mathcal{D}_N$ is used separately again (without serialization) to compute the cross-entropy loss of $h_\theta$, i.e., $\mathcal{L}(\eta, \mathcal{D}_N)$. Accordingly, we re-arrange the training set $\mathcal{D}_N$ to the following serialized/transformed structure for prompting (an example from the Calhousing dataset):

\begin{tcolorbox}[colback=gray!20, boxsep=0mm, left=1.5mm, right=1.5mm, top=2mm, bottom=2mm]
\textbf{Prompt (N=4):}\\
Example 0: The median income is 4.3292. The housing median age is 14.0. The total rooms is 4412.0. The total number of bedrooms is 952.0. The population is 1656.0. The number of households is 874.0. The latitude is 33.77. The longitude is -117.84. Is this house block valuable? Yes or no? The answer is yes. \\

Example 1: The median income is 3.7813. The housing median age is 41.0. The total rooms is 3170.0. The total number of bedrooms is 622.0. The population is 1091.0. The number of households is 528.0. The latitude is 37.9. The longitude is -122.54. Is this house block valuable? Yes or no? The answer is yes. \\

Example 2: The median income is 3.2731. The housing median age is 20.0. The total rooms is 5998.0. The total number of bedrooms is 1320.0. The population is 3185.0. The number of households is 1199.0. The latitude is 33.93. The longitude is -117.45. Is this house block valuable? Yes or no? The answer is no. \\

Example 3: The median income is 1.6955. The housing median age is 24.0. The total rooms is 2316.0. The total number of bedrooms is 599.0. The population is 1829.0. The number of households is 532.0. The latitude is 34.0. The longitude is -117.4. Is this house block valuable? Yes or no? The answer is no. 
\end{tcolorbox}



For computing the cross-entropy loss using $\mathcal{D}_N$, we let \texttt{X\textunderscore train} denote an $N\times d$ tensor which includes the normalized feature values corresponding to the same $N$ examples in the \texttt{prompt}. These feature vectors are used as the input to $h_\theta$ for computing $\mathcal{L}(\eta, \mathcal{D}_N)$. \texttt{y\textunderscore train} denotes the corresponding labels, also used for computing $\mathcal{L}(\eta, \mathcal{D}_N)$. In this example, the datapoints used for creating both the \texttt{prompt} and the \texttt{X/y\textunderscore train} are the same and $N=4$. However, when $N$ is sufficiently large, we may use different subsets of datapoints from $\mathcal{D}_N$ to create the \texttt{prompt} and \texttt{X/y\textunderscore train}. Moreover, if $N$ is even larger, we may generate multiple examples of the above form with non-overlapping subsets from $\mathcal{D}_N$. Notice that when the base model is TabPFN, we directly use $\mathcal{D}_N$ for fine-tuning the transformer model without serialization. 

The few-shot regime poses the inherent problem of \textbf{overfitting}. To overcome this problem, in each epoch, we randomly permute the feature order of $\mathcal{D}_N$. E.g., if in epoch 1 the order was \texttt{(age, education, label)}, in epoch 2 it will be \texttt{(education, age, label)}. All examples in a prompt will have the same feature order. $\mathcal{D}_N$ is chosen to be class-wise balanced. A validation accuracy is computed on the same set  $\mathcal{D}_N$ with a different randomly permuted feature order and is used for determining the hyperparameters such as the number of epochs and the complexity of $h_\theta$. 

See Figure \ref{fig_framework} for an illustration of the framework. Algorithm \ref{algo_tabdistill} summarizes the procedure. See Appendix \ref{appendix_trainingParams} for more details on the exact training parameters corresponding to each dataset and $N$. 

\begin{algorithm}[ht]
\SetAlgoLined
\KwIn{Few-shot dataset $\mathcal{D}_N$, complex model $f$ with encoder $f_E$, transform $g(\cdot)$, number of fine-tuning epochs $T$, number of post-fine-tuning epochs $K$, architecture of the \gls{MLP} $h_\theta$}

\KwOut{Trained \gls{MLP} $h_\theta$}

PHASE 1: Fine-tuning the transformer model to get a good output MLP

Initialize the linear mapping $m_\eta (z):\mathcal{Z} \rightarrow \Theta$ based on the architecture of $h_\theta$ \;

\For{$t \gets 1$ \KwTo $T$}{
    Randomly permute the feature order of $\mathcal{D}_N$ \;
    
    Create subsets $D_s, D_q \subseteq \mathcal{D}_N$\;
    
    Generate $\texttt{Prompt} \leftarrow g(D_s), \texttt{X\textunderscore train} \leftarrow \{x : (x,y)\in D_q\}$ and $\texttt{y\textunderscore train} \leftarrow \{y: (x, y) \in D_q\}$\;
    
    Prompt the base model $f$ and obtain embeddings $z \gets f_E(\texttt{Prompt})$ \;
    
    Infer parameters $\theta \gets m_\eta(z)$ for the MLP $h_\theta$\;
    
    Compute cross-entropy loss of classifier $h_\theta$: $\mathcal{L}(\eta, D_q)$, as in \eqref{eq_trainingLoss} with $x_n \in \texttt{X\textunderscore train}$ and $\; y_n \in \texttt{y\textunderscore train}$\;
    
    Update $\eta$ using gradient descent with gradients $\nabla_\eta \mathcal{L}(\eta, D_q)$
}


Prompt the fine-tuned base model $f$ with original dataset $\mathcal{D}_N$ and obtain the output MLP $h_\theta$\;

PHASE 2: Additional fine-tuning of the obtained MLP if desired





\For{$k \gets 1$ \KwTo $K$}{
    Compute cross-entropy loss $\mathcal{J}(\theta, D_q)$ with $x_n \in \texttt{X\textunderscore train}$ and $\; y_n \in \texttt{y\textunderscore train}$\;
    
    Update $\theta$ using gradient descent with gradients $\nabla_\theta \mathcal{J}(\theta, D_q)$ \;
}

\Return{$h_\theta$} \;
\caption{\methodName{} framework}
\label{algo_tabdistill}
\end{algorithm}

\subsection{Proposed Instantiations of \methodName{} with TabPFN and T0pp}

\textbf{\methodName{} with TabPFN:} TabPFN \cite{hollmann2023TabPFN} is a transformer-based model pre-trained on a large number of synthetic tabular datasets. Tabular data (after pre-processing steps such as normalizing and one-hot encoding) can directly be used as the input to the TabPFN model. Therefore, the transform $g(x)$ in this case is the identity function. The TabPFN library provides a \texttt{scikit-learn}-style \texttt{fit} and \texttt{predict} functionality. In each training epoch, we fit the TabPFN classifier to $\mathcal{D}_N$ (with a randomly-permuted feature order) and obtain $z=f_E(\mathcal{D}_N)$. Next, we get $\theta=m_\eta(z)$ and compute the loss $\mathcal{L}(\eta; \mathcal{D}_N)$ in \eqref{eq_trainingLoss} to perform a gradient descent update on $\eta$. At the end of the training phase, $h_\theta$ is obtained by inputting $\mathcal{D}_N$ to TabPFN encoder without any permutations to the feature order. Finally, $h_\theta$ is fine-tuned on $\mathcal{D}_N$ for additional $K=100$ epochs. The encoder output dimensionality $\text{dim}(\mathcal{Z})$ of TabPFN varies with the number of training examples in multiples of 192. Consequently, the dimensionality of the matrix $A$ in the mapping $m_\eta(z)$ is taken to be $\text{dim}(\Theta) \times 192 N$. 

\textbf{\methodName{} with T0pp:}
The BigScience T0pp \cite{sanh2021t0Models} is an encoder-decoder style \gls{LLM} trained on a large number of English language tasks specified in natural language prompts. This model has been used as the base \gls{LLM} for TabLLM \cite{hegselmann2023TabLLM}. Since the input to the model has to be a natural language prompt, we convert the training data $\mathcal{D}_N$ (or a subset) into natural language using the ``\texttt{The \textless column\textunderscore name> is \textless value>}" style text template. $g(x)$ in this case represents this transform from tabular data to a natural language prompt. See Figure \ref{fig_transformerClassifiers} for a detailed illustration. Appendix \ref{appendix_t0ppPrompts} lists example serializations used for each dataset. The training and inference phases are similar to that of \methodName{} with TabPFN. In the end, the resultant \gls{MLP} $h_\theta$ is fine-tuned on $\mathcal{D}_N$ for additional $K=100$ epochs. The dimensionality of the encoder output $z=f_E(g(\mathcal{D}_N))$ is 4096. Therefore, the dimensionality of the matrix $A$ in the mapping $m_\eta(z)$ is taken to be $\text{dim}(\Theta) \times 4096$.

\section{Experimental Results}
\label{sec_experiments}
\textbf{Datasets and metrics:} We evaluate the \methodName{} framework on five publicly available tabular datasets: \textit{Bank} (UCI Bank Marketing) \cite{bankDataset}, \textit{Blood} (UCI Blood Transfusion Service Center) \cite{bloodDataset}, \textit{Calhousing} (California Housing Prices) \cite{calhousingDataset}, \textit{Heart} (UCI Heart Disease) \cite{heartDataset} and \textit{Income} (Census Income) \cite{incomeDataset}. We divide each dataset into a train and a test split. $\mathcal{D}_N$ is selected from the training split. More details about the datasets are given in Table \ref{tab_datasets}. Performance of all the models is compared with respect to the \gls{ROC-AUC} metric. We consider the few-shot regime where the number of training examples is very low, specifically, $N \in \{4, 8, 16, 32, 64\}$.


\textbf{Baselines and $h_\theta$:} The architecture of $h_\theta$ is constant across all the experiments, unless specified explicitly. $h_\theta$ consisted of two hidden layers (hence, four layers in total, i.e., $R=4$) with 10 neurons each (i.e., $L=10$). We compare \methodName{} with 3 simple and efficient classical baselines: logistic regression, XGBoost \cite{chen2016xgboost}, an \gls{MLP} with an architecture similar to $h_\theta$ but trained independently. In addition, we provide a performance comparison w.r.t. the base models TabPFN \cite{hollmann2023TabPFN} and T0pp \cite{hegselmann2023TabLLM} for completeness. All the models use the same set of labeled examples as \methodName{} for training. Logistic regression and XGBoost are the best performing classical models in \cite{hegselmann2023TabLLM}, and hence, provide a strong baseline. The performance of the independently trained \gls{MLP} helps observing the performance improvement obtained as a result of the distillation process.

\textbf{Hyperparameters:} Hyperparameters of all the baselines except were tuned using 4-fold cross-validation similar to \cite{hegselmann2023TabLLM}, except in the case of training set size 4. When the training set size is 4, 2-fold cross-validation was used. We use \texttt{Scikit-learn}'s \texttt{GridSearchCV} and \texttt{RandomizedSearchCV} for tuning the hyperparameters. For XGBoost and \gls{MLP}, we adopt the hyperparameter search ranges given in \cite{grinsztajn2022WhyDoTreeBasedModels}. However, we keep the architecture of the \gls{MLP} fixed to that of $h_\theta$. See Table \ref{tab_hpo} for more details on hyperparameter tuning of the baselines. TabPFN does not require any hyperparameters to be tuned \cite{hollmann2023TabPFN}. \texttt{Weights and Biases} sweeps were used for optimizing the hyperparameters of \methodName{}, based on a validation score computed using the same training set $\mathcal{D}_N$. See Figure \ref{fig_hpoCalhousing} in Appendix \ref{appendix_trainingParams} for an example sweep.





\textbf{Main observations:} Table \ref{tab_AUCPerformanceBaselines} presents the \gls{ROC-AUC} of \methodName{} along with that of the baselines, over the five tabular datasets. \methodName{} shows superior performance over its classical counterparts particularly in the very few-shot regime. In general, the performance increases with the number of labeled examples available (i.e., with increasing $N$). Out of the three classical baselines, none seems to be universally better in performance across the datasets or the number of labeled examples. \methodName{} + TabPFN shows better performance that \methodName{}+T0pp in most cases, except in the Income dataset, where \methodName{}+T0pp performs consistently better.

\begin{table}[ht]
\caption{Test ROC-AUC performance of \methodName{} compared with the classical baselines. Best performance corresponding to each $N$ and dataset is emphasized in \textbf{bold}. Reported values are the average of 5 runs with different random states. The standard deviations are given as subscripts.}
\label{tab_AUCPerformanceBaselines}
\begin{center}
\begin{tabular}{llccccc}
\toprule
\multirow{2}{*}{\bf Dataset}  &\multirow{2}{*}{\bf Method} &\multicolumn{5}{c}{\bf Number of labeled examples ($N$)} \\
\cmidrule(lr){3-7}
& & 4 & 8 & 16 & 32 & 64\\ 
\midrule

\multirow{5}{*}{Bank} & MLP                           & $0.57_{.08}$             & $0.61_{.11}$             & $\mathbf{0.72_{.05}}$    & $0.76_{.04}$             & $\mathbf{0.81_{.03}}$\\
                      & Logistic Regression           & $0.54_{.10}$             & $0.65_{.06}$             & $\mathbf{0.72_{.03}}$    & $0.73_{.04}$             & $0.77_{.04}$\\
                      & XGBoost                       & $0.50_{.00}$             & $0.56_{.10}$             & $\mathbf{0.72_{.08}}$    & $0.78_{.04}$             & $\mathbf{0.81_{.02}}$\\
                      & \methodName{} + TabPFN (ours) & $\mathbf{0.72_{.01}}$    & $\mathbf{0.67_{.06}}$    & $0.68_{.02}$             & $\mathbf{0.79_{.02}}$    & $\mathbf{0.81_{.02}}$\\
                      & \methodName{} + T0pp (ours)   & $0.70_{.02}$             & $\mathbf{0.67_{.02}}$    & $\mathbf{0.72_{.01}}$    & $0.74_{.02}$             & $0.80_{.02}$\\
\midrule

\multirow{5}{*}{Blood} & MLP                           & $0.57_{.10}$             & $0.61_{.09}$             & $0.60_{.09}$             & $0.61_{.07}$             & $0.67_{.08}$\\
                       & Logistic Regression           & $0.60_{.16}$             & $0.66_{.12}$             & $0.63_{.11}$             & $0.65_{.10}$             & $0.73_{.03}$\\
                       & XGBoost                       & $0.50_{.00}$             & $0.55_{.09}$             & $0.55_{.07}$             & $0.65_{.07}$             & $0.72_{.02}$\\
                       & \methodName{} + TabPFN (ours) & $0.56_{.07}$             & $\mathbf{0.67_{.05}}$    & $\mathbf{0.69_{.07}}$    & $\mathbf{0.68_{.09}}$    & $\mathbf{0.75_{.00}}$\\
                       & \methodName{} + T0pp (ours)   & $\mathbf{0.62_{.08}}$    & $0.58_{.08}$             & $0.67_{.06}$             & $0.67_{.04}$             & $0.68_{.06}$\\
\midrule

\multirow{5}{*}{Calhousing} & MLP                           & $0.49_{.07}$             & $0.63_{.10}$             & $0.72_{0.12}$            & $0.79_{.07}$             & $0.82_{.04}$\\
                            & Logistic Regression           & $0.59_{.10}$             & $0.66_{.13}$             & $0.74_{.14}$             & $\mathbf{0.83_{.04}}$    & $\mathbf{0.89_{.01}}$\\
                            & XGBoost                       & $0.50_{.00}$             & $0.57_{.10}$             & $\mathbf{0.75_{.04}}$    & $0.75_{.06}$             & $0.81_{.06}$\\
                            & \methodName{} + TabPFN (ours) & $0.64_{.06}$             & $0.65_{.03}$             & $0.65_{.03}$             & $0.77_{.03}$             & $0.84_{.00}$\\
                            & \methodName{} + T0pp (ours)   & $\mathbf{0.67_{.05}}$    & $\mathbf{0.67_{.03}}$    & $0.66_{.05}$             & $0.74_{.03}$             & $0.81_{.01}$\\
\midrule

\multirow{5}{*}{Heart}      & MLP                           & $0.63_{.14}$             & $0.69_{.17}$             & $0.83_{.08}$             & $0.84_{.05}$             & $0.84_{.03}$\\
                            & Logistic Regression           & $0.67_{.28}$             & $0.84_{.07}$             & $\mathbf{0.90_{.03}}$    & $\mathbf{0.90_{.01}}$    & $\mathbf{0.91_{.02}}$\\
                            & XGBoost                       & $0.50_{.00}$             & $0.60_{.13}$             & $0.84_{.07}$             & $0.84_{.07}$             & $0.90_{.02}$\\
                            & \methodName{} + TabPFN (ours) & $\mathbf{0.77_{.10}}$    & $\mathbf{0.85_{.01}}$    & $0.81_{.05}$             & $0.85_{.03}$             & $0.87_{.02}$\\
                            & \methodName{} + T0pp (ours)   & $0.76_{.11}$             & $0.76_{.09}$             & $0.82_{.05}$             & $0.88_{.01}$             & $0.90_{.01}$\\
\midrule

\multirow{5}{*}{Income} & MLP                           & $0.51_{.10}$             & $0.69_{.05}$              & $0.74_{.07}$             & $0.78_{.04}$             & $0.79_{.04}$\\
                        & Logistic Regression           & $\mathbf{0.76_{.07}}$    & $0.75_{.09}$              & $0.79_{.02}$             & $0.82_{.02}$             & $0.84_{.03}$\\
                        & XGBoost                       & $0.50_{.00}$             & $0.57_{.11}$              & $0.65_{.14}$             & $0.80_{.02}$             & $0.81_{.01}$\\
                        & \methodName{} + TabPFN (ours) & $0.68_{.08}$             & $0.75_{.03}$              & $0.80_{.02}$             & $0.81_{.02}$             & $0.83_{.01}$\\
                        & \methodName{} + T0pp (ours)   & $0.70_{.03}$             & $\mathbf{0.77_{.02}}$     & $\mathbf{0.83_{.01}}$    & $\mathbf{0.83_{.02}}$    & $\mathbf{0.85_{.01}}$\\

\bottomrule
\end{tabular}
\end{center}
\end{table}

\begin{figure}[h]
    \centering
    \begin{subfigure}[h]{0.45\textwidth}
        \centering
        \includegraphics[width=\textwidth]{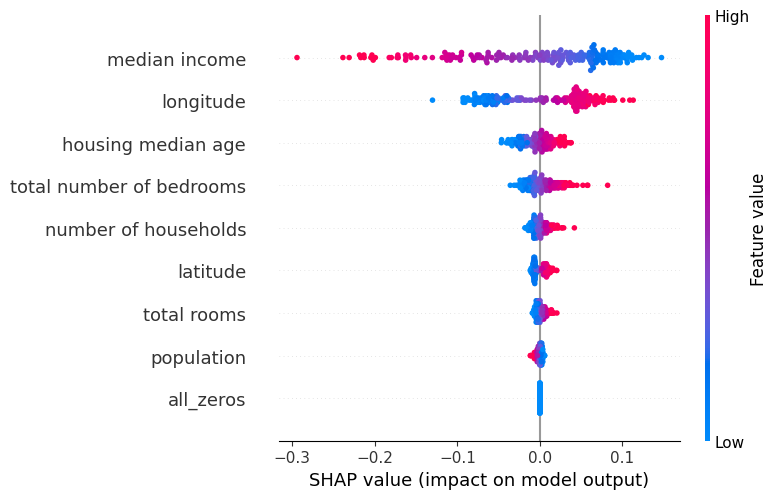} 
        \caption{Logistic regression trained on $\mathcal{D}_N$}
        \label{fig_shapLR}
    \end{subfigure}
    \begin{subfigure}[h]{0.45\textwidth}
        \centering
        \includegraphics[width=\textwidth]{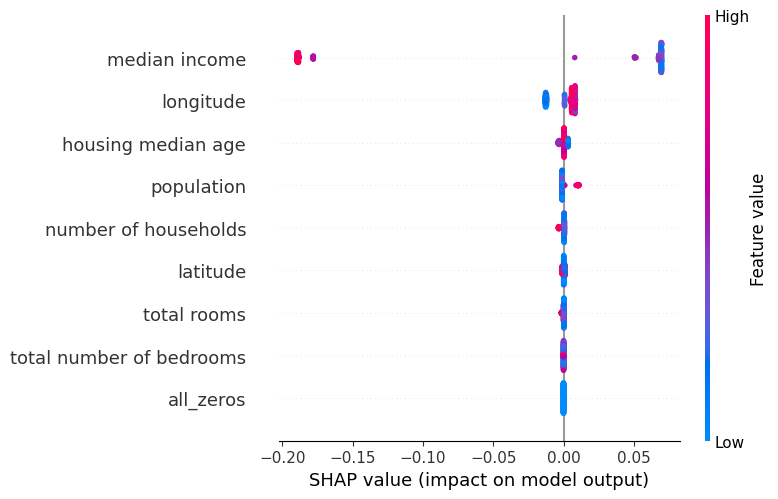}
        \caption{XGBoost trained on $\mathcal{D}_N$}
        \label{fig_shapXGBoost}
    \end{subfigure}
    \begin{subfigure}[h]{0.45\textwidth}
        \centering
        \includegraphics[width=\textwidth]{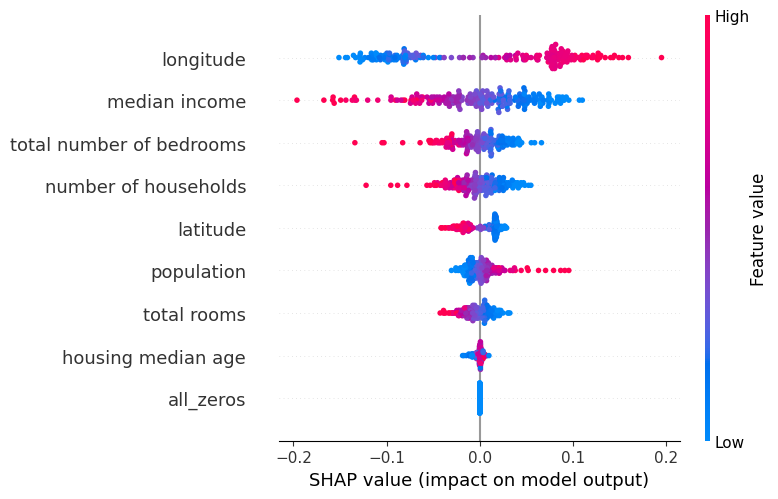}
        \caption{$h_\theta$ distilled on $\mathcal{D}_N$}
        \label{fig_shapDistilledMLP}
    \end{subfigure}
    \begin{subfigure}[h]{0.45\textwidth}
        \centering
        \includegraphics[width=\textwidth]{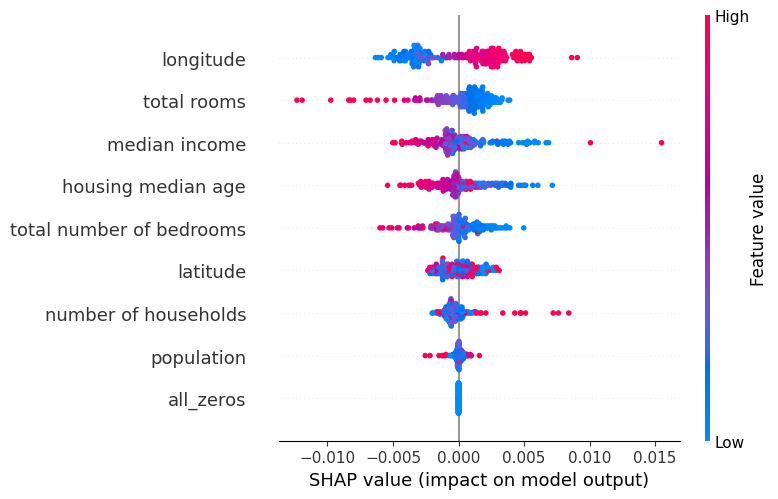}
        \caption{$h_\theta$ distilled on feature-permuted $\mathcal{D}_N$}
        \label{fig_shapPermutedMLP}
    \end{subfigure}
    \caption{\texttt{SHAP} feature attributions. Computed on the Calhousing dataset with TabPFN as the base model $f$. Training set size $N$ is 16. 200 samples were used for computing the beeswarm plots.}
    \label{fig_shapBaselines}
\end{figure}

\textbf{Effect of the complexity of $h_\theta$:}  In Table \ref{tab_AUCPerformanceMLPDepth} we study the effect of the complexity of $h_\theta$ measured in terms of the number of layers $R$. The layer size $L$ is kept constant at 10. Bank dataset and TabPFN base model were used for the evaluation. As it is evident from the results, when the complexity of $h_\theta$ increases beyond a certain limit, the performance degrades.

\textbf{Performance with respect to the base models:}
Table \ref{tab_baseModelPerformance} presents the performance of the \gls{MLP} $h_\theta$ obtained using \methodName{} compared to the corresponding transformer-based model $f$ used for distillation. Interestingly, in some cases, the \gls{MLP} $h_\theta$ distilled using our method surpasses the performance of the base model $f$ which it was distilled from.

\textbf{Feature attribution comparison:} We compute the \texttt{SHAP} feature attribution scores \cite{lundberg2017unified,shapley1953shapleyValue} using the \texttt{SHAP} library for the classical baseline models logistic regression and XGBoost, and $h_\theta$ using the Calhousing dataset. The number of training examples used was 16 and the base model $f$ was TabPFN. Figure \ref{fig_shapBaselines} shows the corresponding beeswarm plots for each baseline. We observe that the \texttt{median income} and the \texttt{longitude} have a greater impact on the output across all the models, indicating that the distilled models are consistent with the baselines trained in the ordinary fashion (See figures \ref{fig_shapLR}, \ref{fig_shapXGBoost} and \ref{fig_shapDistilledMLP}). We also compute the attributions scores corresponding to an \gls{MLP} distilled using a $\mathcal{D}_N$ with feature columns permuted (Figure \ref{fig_shapPermutedMLP}). Interestingly, despite the permutation, our distilled model displays feature importances similar to the original $h_\theta$. Hence, it is evident  that the base model has correctly identified the correlation between the \gls{MLP} weights and the feature order.

\section{Conclusion}
We introduce \methodName{}, a novel distillation framework for extracting the pre-trained knowledge of transformer models into neural networks for classifying tabular data. The framework produces \glspl{MLP} with enhanced performance particularly when the labeled data is limited. Experiments show that the resulting \glspl{MLP} surpass the classical machine learning models such as XGBoost and logistic regression, and in some cases, the initial transformer model used for distilling itself in the few-shot regime. \methodName{} can be used to generate scalable, computationally efficient models with a small number of training data, bringing together the advantages of transformers and classical models.

\begin{table}[h!]
\caption{Test ROC-AUC performance of \methodName{} with different MLP complexities}
\label{tab_AUCPerformanceMLPDepth}
\begin{center}
\begin{tabular}{ccccc}
\toprule
\multirow{2}{2cm}{\bf \# Labeled examples} & \multicolumn{4}{c}{\bf Number of layers (R)} \\
\cmidrule{2-5}
& 2 & 4 & 8 & 16 \\
\midrule
4 & $0.72_{.02}$ & $0.72_{.01}$ & $0.65_{.09}$ & $0.53_{.04}$\\
8 & $0.74_{.01}$ & $0.67_{.06}$ & $0.72_{.03}$ & $0.50_{.00}$\\
\bottomrule
\end{tabular}
\end{center}
\end{table}

\textbf{Limitations and future directions:} While \methodName{} produces \glspl{MLP} which surpass the classical models in the few-shot regime, the performance gain is limited when the number of labeled examples increase. Hence, there is room for improvement when the training set is large. The linear mapping function $m_\eta(\cdot)$ used in the current experiments can be replaced with other alternatives to potentially achieve performance improvements. Moreover, the extracted \gls{MLP} may inherit the biases of the base model, although it can be mitigated to some extent through the \gls{MLP} finetuning in the second phase. Future work will explore extensions beyond tabular classification, e.g., natural language inference~\cite{hamman2025few} or instruction tuning~\cite{fu2025t} in few-shot settings.

\paragraph{Reproducibility Statement:}
The \methodName{} framework has been explained in detail in Section \ref{sec_methodology}. Details on the experimental setup including the datasets and the baselines are given in Section \ref{sec_experiments}. Further details including hyperparameters and the training setup are given in Appendix \ref{appendix_trainingParams}. All the experiments were done on a computer with a 3.5 GHz AMD EPYC 7763 64-Core Processor and an Nvidia RTX 6000 Ada GPU.

\begin{table}[h]
\caption{Test ROC-AUC performance of \methodName{} compared with the base model $f$. Best performance corresponding to each $N$ and dataset is emphasized in \textbf{bold}. Reported values are the average of 5 runs with different random states. The standard deviations are given as subscripts.}
\label{tab_baseModelPerformance}
\begin{center}
\begin{tabular}{p{1.4cm}lccccc}
\toprule
\multirow{2}{*}{\bf Dataset}  &\multirow{2}{*}{\bf Method} &\multicolumn{5}{c}{\bf Number of labeled examples ($N$)} \\
\cmidrule(lr){3-7}
& & 4 & 8 & 16 & 32 & 64\\ 
\midrule
\multirow{4}{*}{Bank} & TabPFN (11M params)                                  & $0.62_{.05}$             & $\mathbf{0.68_{.08}}$     & $\mathbf{0.75_{.08}}$    & $\mathbf{0.82_{.05}}$    & $\mathbf{0.86_{.02}}$\\
                      & \methodName{} + TabPFN (ours)                        & $\mathbf{0.72_{.01}}$    & $0.67_{.06}$              & $0.68_{.02}$             & $0.79_{.02}$             & $0.81_{.02}$\\
                      \cmidrule{2-7}
                      & T0pp (TabLLM, 11B params)\textsuperscript{$\dagger$} & $0.59_{.10}$             & $0.64_{.05}$              & $0.65_{.05}$             & $0.64_{0.6}$             & $0.69_{.03}$\\
                      & \methodName{} + T0pp (ours)                          & $\mathbf{0.70_{.02}}$    & $\mathbf{0.67_{.02}}$     & $\mathbf{0.72_{.01}}$    & $\mathbf{0.74_{.02}}$    & $\mathbf{0.80_{.02}}$\\
\midrule

\multirow{4}{*}{Blood} & TabPFN (11M params)                                  & $0.55_{.20}$             & $0.61_{.14}$             & $0.59_{.12}$             & $\mathbf{0.68_{.07}}$     & $0.73_{.02}$\\
                       & \methodName{} + TabPFN (ours)                        & $\mathbf{0.56_{.07}}$    & $\mathbf{0.67_{.05}}$    & $\mathbf{0.69_{.07}}$    & $\mathbf{0.68_{.09}}$     & $\mathbf{0.75_{.00}}$\\
                       \cmidrule{2-7}
                       & T0pp (TabLLM, 11B params)\textsuperscript{$\dagger$} & $0.58_{.09}$              & $\mathbf{0.66_{.03}}$   & $0.66_{.07}$             & $\mathbf{0.68_{.04}}$     & $\mathbf{0.68_{.04}}$\\
                       & \methodName{} + T0pp (ours)                          & $\mathbf{0.62_{.08}}$     & $0.58_{.08}$            & $\mathbf{0.67_{.06}}$    & $0.67_{.04}$              & $\mathbf{0.68_{.06}}$\\
\midrule

\multirow{4}{1cm}{\small Calhousing} & TabPFN (11M params)                         & $0.59_{.08}$           & $\mathbf{0.70_{.10}}$   & $\mathbf{0.83_{.04}}$  & $\mathbf{0.84_{.04}}$  & $\mathbf{0.88_{.02}}$\\
                            & \methodName{} + TabPFN (ours)                        & $\mathbf{0.64_{.06}}$  & $0.65_{.03}$            & $0.65_{.03}$           & $0.77_{.03}$           & $0.84_{.00}$\\
                            \cmidrule{2-7}
                            & T0pp (TabLLM, 11B params)\textsuperscript{$\dagger$} & $0.63_{.05}$           & $0.60_{.07}$            & $\mathbf{0.70_{.08}}$  & $\mathbf{0.77_{.08}}$  & $0.77_{0.4}$\\
                            & \methodName{} + T0pp (ours)                          & $\mathbf{0.67_{.05}}$  & $\mathbf{0.67_{.03}}$   & $0.66_{.05}$           & $0.74_{.03}$           & $\mathbf{0.81_{.01}}$\\
\midrule

\multirow{4}{1cm}{\small Heart} & TabPFN (11M params)                              & $0.67_{.21}$           & $0.80_{.08}$            & $\mathbf{0.89_{.02}}$  & $\mathbf{0.88_{.01}}$  & $\mathbf{0.91_{.02}}$\\
                            & \methodName{} + TabPFN (ours)                        & $\mathbf{0.77_{.10}}$  & $\mathbf{0.85_{.01}}$   & $0.81_{.05}$           & $0.85_{.03}$           & $0.87_{.02}$\\
                            \cmidrule{2-7}
                            & T0pp (TabLLM, 11B params)\textsuperscript{$\dagger$} & $\mathbf{0.76_{.14}}$  & $\mathbf{0.83_{.05}}$   & $\mathbf{0.87_{.04}}$  & $0.87_{.06}$           & $\mathbf{0.91_{.01}}$\\
                            & \methodName{} + T0pp (ours)                          & $\mathbf{0.76_{.11}}$  & $0.76_{.09}$            & $0.82_{.05}$           & $\mathbf{0.88_{.01}}$  & $0.90_{.01}$\\
\midrule

\multirow{4}{*}{Income} & TabPFN (11M params)                                  & $\mathbf{0.69_{.06}}$    & $0.74_{.09}$           & $0.78_{.01}$             & $\mathbf{0.82_{.03}}$  & $\mathbf{0.84_{.01}}$\\
                        & \methodName{} + TabPFN (ours)                        & $0.68_{.08}$             & $\mathbf{0.75_{.03}}$  & $\mathbf{0.80_{.02}}$    & $0.81_{.02}$           & $0.83_{.01}$\\
                        \cmidrule{2-7}
                        & T0pp (TabLLM, 11B params)\textsuperscript{$\dagger$} & $\mathbf{0.84_{.01}}$    & $\mathbf{0.84_{.02}}$  & $\mathbf{0.84_{.04}}$    & $\mathbf{0.84_{.01}}$  & $0.84_{.02}$\\
                        & \methodName{} + T0pp (ours)                          & $0.70_{.03}$             & $0.77_{.02}$           & $0.83_{.01}$             & $0.83_{.02}$           & $\mathbf{0.85_{.01}}$\\
\bottomrule
\end{tabular}
\end{center}
\end{table}

\FloatBarrier
\renewcommand\bibname{\large References}
\printbibliography

\newpage
\appendix
\section{Training setup and hyperparameter optimization}
\label{appendix_trainingParams}
The PHASE 1 fine-tuning was carried out for 300 epochs. The epoch with the best validation accuracy (computed using a randomly permuted version of the $\mathcal{D}_N$) was used for inferring the weights for the final $h_\theta$. Learning rates were selected from the set $[1e^{-6}, 2e^{-4}]$. Adam optimizer was used with a weight decay of $1e^{-3}$. For some of the experiments with $N=4$, weight decay was set to $0$. The subsets $D_s$ and $D_q$ of the training set $\mathcal{D}_N$ were selected as per Table \ref{tab_datasetPartition}. More details on the datasets including the test-train split sizes is given in Table \ref{tab_datasets}.

\begin{table}[ht]
\caption{The scheme of partitioning $\mathcal{D}_N$}
\label{tab_datasetPartition}
\begin{center}
\begin{tabular}{cccccc}
\toprule
\multirow{2}{*}{\bf Parameter} & \multicolumn{5}{c}{\bf Number of labeled examples} \\
\cmidrule{2-6}
& 4 & 8 & 16 & 32 & 64\\
\midrule
$|D_s|$ & 4 & 4 & 8 & 8 & 8\\ 
$|D_q|$ & 4 & 4 & 8 & 8 & 8\\
$D_s=D_q$? & True & True & False & False & False \\
\# of ($D_s, D_q$) pairs & 1 & 2 & 1 & 2 & 4 \\
\bottomrule
\end{tabular}
\end{center}
\end{table}

\begin{table}[ht]
\caption{Dataset details}
\label{tab_datasets}
\begin{center}
\begin{tabular}{ccccL{5cm}}
\toprule
\textbf{Dataset} & \textbf{\# Features} & \textbf{Test size} & \textbf{Train size} & \textbf{Target} \\
\midrule
Bank & 16 & 43211 & 2000 & To predict whether the client will subscribe a term deposit\\
Blood & 4 & 374 & 374 & To predict whether a person would donate blood\\
Calhousing & 12 & 19640 & 1000 & To predict whether a given house block is valuable or not\\
Heart & 13 & 459 & 459 & To predict whether a patient has a heart disease \\
Income & 12 & 44222 & 1000 & To predict whether a person's annual income exceeds 50K\\
\bottomrule
\end{tabular}
\end{center}
\end{table}

\begin{table}[h]
\caption{Hyperparameter ranges used for classical baseline models}
\label{tab_hpo}
\begin{center}
\begin{tabular}{clll}
\toprule
\textbf{Model} & \textbf{Hyperparameter} & \textbf{Range/Distribution} & \textbf{Method} \\
\midrule
\multirow{4}{*}{\gls{MLP}} & Number of layers & 4 & \multirow{4}{*}{Grid search} \\
& Hidden layer size & 10 & \\
& Number of epochs & [30, 50, 100, 300] & \\
& Learning rate & [1e-5, 1e-4, 1e-3, 1e-2] & \\
\midrule

Logistic Regression & C & [0.01, 0.1, 1, 10] & Grid search\\
\midrule

\multirow{10}{*}{XGBoost} & Max depth & UniformInt[1,11] & \multirow{10}{2.5cm}{Randomized search with 20 iterations}\\
& Number of estimators & 1000 & \\
& Min child weight & LogUniformInt[1, 1e2] & \\
& Subsample & Uniform[0.5, 1] & \\
& Learning rate & LogUniform[1e-5, 0.7] & \\
& Column sample by level & Uniform[0.5, 1] & \\
& Column sample by tree & Uniform[0.5, 1] & \\
& Gamma & LogUniform[1e-8, 7] & \\
& Lambda & LogUniform[1, 4] & \\
& Alpha & LogUniform[1e-8, 1e2] & \\
\bottomrule
\end{tabular}
\end{center}
\end{table}

\begin{figure}[h!]
    \centering
    \includegraphics[width=0.9\linewidth]{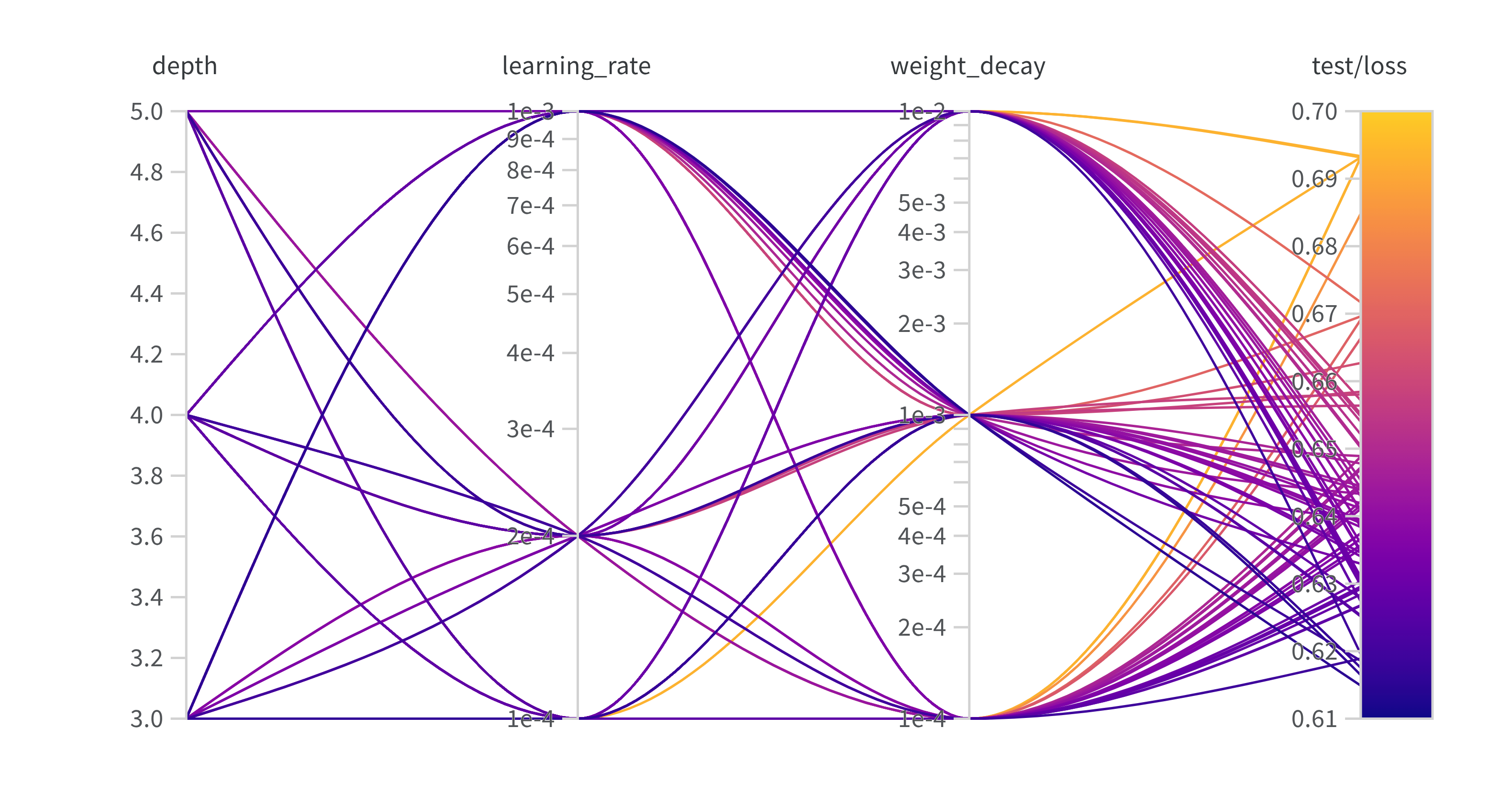}
    \caption{\texttt{Weights and Biases} sweeps used for optimizing hyperparameters for \methodName{} with TabPFN and Calhousing dataset, 64 training examples. Each column (y-axes) represents a tunable hyperparameter and its value. The right-most column represents the loss over a validation set. Each colored line traversing from left to right represents a set of values selected for each hyperparameter (i.e., a single run of the sweep). The color of the line corresponds to the validation loss achieved by this set of hyperparameters.}
    \label{fig_hpoCalhousing}
\end{figure}

Hyperparameter optimization for the classical baseline models was done using \texttt{scikit-learn}'s \texttt{GridSearchCV} and \texttt{RandomizedSearchCV} methods. The search ranges are given in Table \ref{tab_hpo}. Hyperparameters for the \methodName{} framework were deteremined using \texttt{Weights and Biases} sweeps. Figure \ref{fig_hpoCalhousing} illustrates one such sweep corresponding to the TabPFN base model and the Calhousing dataset.

\FloatBarrier
\section{Serializations used for prompting T0pp}
\label{appendix_t0ppPrompts}

\begin{tcolorbox}[breakable, enhanced, colback=gray!20, boxsep=0mm, left=2mm, right=2mm, top=2mm, bottom=2mm]
\textbf{Bank dataset, N=4:}\\
Example 0: The age is 29.0. The job is blue-collar. The marital status is married. The education is secondary. The default is no. The account balance is 314.0. The housing loan is available. The personal loan is available. The contact communication type is cellular. The last contact day of the month is 17.0. The last contact month of year is apr. The last contact duration, in seconds is 357.0. The number of contacts in campaign is 1.0. The days since last contact is -1.0. The number of previous contacts is 0.0. The previous contact outcome is unknown. Does this client subscribe to a term deposit? Yes or no? The answer is no.\\

Example 1: The age is 62.0. The job is housemaid. The marital status is married. The education is unknown. The default is no. The account balance is 2021.0. The housing loan is not available. The personal loan is not available. The contact communication type is telephone. The last contact day of the month is 26.0. The last contact month of year is feb. The last contact duration, in seconds is 361.0. The number of contacts in campaign is 1.0. The days since last contact is -1.0. The number of previous contacts is 0.0. The previous contact outcome is unknown. Does this client subscribe to a term deposit? Yes or no? The answer is yes.\\

Example 2: The age is 32.0. The job is blue-collar. The marital status is single. The education is secondary. The default is no. The account balance is 3.0. The housing loan is available. The personal loan is not available. The contact communication type is unknown. The last contact day of the month is 23.0. The last contact month of year is may. The last contact duration, in seconds is 108.0. The number of contacts in campaign is 3.0. The days since last contact is -1.0. The number of previous contacts is 0.0. The previous contact outcome is unknown. Does this client subscribe to a term deposit? Yes or no? The answer is no.\\

Example 3: The age is 36.0. The job is management. The marital status is married. The education is tertiary. The default is no. The account balance is 203.0. The housing loan is not available. The personal loan is not available. The contact communication type is cellular. The last contact day of the month is 25.0. The last contact month of year is jan. The last contact duration, in seconds is 255.0. The number of contacts in campaign is 1.0. The days since last contact is 88.0. The number of previous contacts is 1.0. The previous contact outcome is success. Does this client subscribe to a term deposit? Yes or no? The answer is yes.

\end{tcolorbox}

\begin{tcolorbox}[breakable, enhanced, colback=gray!20, boxsep=0mm, left=2mm, right=2mm, top=2mm, bottom=2mm]
\textbf{Blood dataset, N=4:}\\
Example 0: The previous blood donation record of a person is as follows: The months since last donation is 14.0. The total number of donations is 3.0. The total amount of blood donated (in cc) is 750.0. The months since first donation is 21.0. Will this person donate blood next time? Yes or no? The answer is no.\\

Example 1: The previous blood donation record of a person is as follows: The months since last donation is 4.0. The total number of donations is 2.0. The total amount of blood donated (in cc) is 500.0. The months since first donation is 4.0. Will this person donate blood next time? Yes or no? The answer is yes.\\

Example 2: The previous blood donation record of a person is as follows: The months since last donation is 16.0. The total number of donations is 7.0. The total amount of blood donated (in cc) is 1750.0. The months since first donation is 87.0. Will this person donate blood next time? Yes or no? The answer is yes.\\

Example 3: The previous blood donation record of a person is as follows: The months since last donation is 11.0. The total number of donations is 5.0. The total amount of blood donated (in cc) is 1250.0. The months since first donation is 35.0. Will this person donate blood next time? Yes or no? The answer is no.
\end{tcolorbox}

\begin{tcolorbox}[breakable, enhanced, colback=gray!20, boxsep=0mm, left=2mm, right=2mm, top=2mm, bottom=2mm]
\textbf{Calhousing dataset, N=4:}\\
Example 0: The median income is 4.3292. The housing median age is 14.0. The total rooms is 4412.0. The total number of bedrooms is 952.0. The population is 1656.0. The number of households is 874.0. The latitude is 33.77. The longitude is -117.84. Is this house block valuable? Yes or no? The answer is yes. \\

Example 1: The median income is 3.7813. The housing median age is 41.0. The total rooms is 3170.0. The total number of bedrooms is 622.0. The population is 1091.0. The number of households is 528.0. The latitude is 37.9. The longitude is -122.54. Is this house block valuable? Yes or no? The answer is yes. \\

Example 2: The median income is 3.2731. The housing median age is 20.0. The total rooms is 5998.0. The total number of bedrooms is 1320.0. The population is 3185.0. The number of households is 1199.0. The latitude is 33.93. The longitude is -117.45. Is this house block valuable? Yes or no? The answer is no. \\

Example 3: The median income is 1.6955. The housing median age is 24.0. The total rooms is 2316.0. The total number of bedrooms is 599.0. The population is 1829.0. The number of households is 532.0. The latitude is 34.0. The longitude is -117.4. Is this house block valuable? Yes or no? The answer is no. 
\end{tcolorbox}

\begin{tcolorbox}[breakable, enhanced, colback=gray!20, boxsep=0mm, left=2mm, right=2mm, top=2mm, bottom=2mm]
\textbf{Heart dataset, N=4:}\\
Example 0: The age is 49. The sex is male. The chest pain type is asymptomatic. The resting blood pressure is 140. The cholesterol is 185. The fasting blood sugar is below 120 mg per dl. The resting ECG is normal. The maximum heart rate is 130. The occurence of exercise induced angina is no. The oldpeak is 0.0. The ST slope is up. Does this patient have a heart disease? Yes or no? The answer is no.\\

Example 1: The age is 62. The sex is male. The chest pain type is asymptomatic. The resting blood pressure is 158. The cholesterol is 210. The fasting blood sugar is above 120 mg per dl. The resting ECG is normal. The maximum heart rate is 112. The occurence of exercise induced angina is yes. The oldpeak is 3.0. The ST slope is down. Does this patient have a heart disease? Yes or no? The answer is yes.\\

Example 2: The age is 61. The sex is male. The chest pain type is asymptomatic. The resting blood pressure is 130. The cholesterol is 0. The fasting blood sugar is above 120 mg per dl. The resting ECG is normal. The maximum heart rate is 77. The occurence of exercise induced angina is no. The oldpeak is 2.5. The ST slope is flat. Does this patient have a heart disease? Yes or no? The answer is yes.\\

Example 3: The age is 61. The sex is male. The chest pain type is typical angina. The resting blood pressure is 134. The cholesterol is 234. The fasting blood sugar is below 120 mg per dl. The resting ECG is normal. The maximum heart rate is 145. The occurence of exercise induced angina is no. The oldpeak is 2.6. The ST slope is flat. Does this patient have a heart disease? Yes or no? The answer is yes.
\end{tcolorbox}

\begin{tcolorbox}[breakable, enhanced, colback=gray!20, boxsep=0mm, left=2mm, right=2mm, top=2mm, bottom=2mm]
\textbf{Income dataset, N=4:}\\
Example 0: The age is 35. The workclass is Private. The education is HS-grad. The marital-status is Married-civ-spouse. The occupation is Transport-moving. The relationship is Husband. The race is White. The sex is Male. The capital-gain is 0. The capital-loss is 0. The hours-per-week is 30. The native-country is United-States. Does this person make over 50K a year? Answer with Yes or No. The answer is No.\\

Example 1: The age is 32. The workclass is Self-emp-not-inc. The education is 10th. The marital-status is Married-civ-spouse. The occupation is Exec-managerial. The relationship is Husband. The race is White. The sex is Male. The capital-gain is 0. The capital-loss is 0. The hours-per-week is 55. The native-country is United-States. Does this person make over 50K a year? Answer with Yes or No. The answer is No.\\

Example 2: The age is 56. The workclass is Private. The education is Some-college. The marital-status is Married-civ-spouse. The occupation is Sales. The relationship is Husband. The race is White. The sex is Male. The capital-gain is 0. The capital-loss is 0. The hours-per-week is 45. The native-country is United-States. Does this person make over 50K a year? Answer with Yes or No. The answer is Yes.\\

Example 3: The age is 44. The workclass is State-gov. The education is Masters. The marital-status is Married-civ-spouse. The occupation is Prof-specialty. The relationship is Husband. The race is White. The sex is Male. The capital-gain is 7688. The capital-loss is 0. The hours-per-week is 50. The native-country is United-States. Does this person make over 50K a year? Answer with Yes or No. The answer is Yes.

\end{tcolorbox}

\end{document}